# Host-based anomaly detection using Eigentraces feature extraction and one-class classification on system call trace data

**Ehsan Aghaei [1], Gursel Serpen[2]**

[1] Electrical Engineering and Computer Science, University of Toledo
2801 Bancroft St, Toledo, OH 43606, USA
*eaghaei@uncc.edu*

[2] Electrical Engineering and Computer Science, University of Toledo
2801 Bancroft St, Toledo, OH 43606, USA
*gursel.serpen@utoledo.edu*

*Abstract*: **This paper proposes a methodology for host-based anomaly detection using a semi-supervised algorithm namely one-class classifier combined with a PCA-based feature extraction technique called Eigentraces on system call trace data. The one-class classification is based on generating a set of artificial data using a reference distribution and combining the target class probability function with artificial class density function to estimate the target class density function through the Bayes formulation. The benchmark dataset, ADFA-LD, is employed for the simulation study. ADFA-LD dataset contains thousands of system call traces collected during various normal and attack processes for the Linux operating system environment. In order to pre-process and to extract features, windowing on the system call trace data followed by the principal component analysis which is named as Eigentraces is implemented. The target class probability function is modeled separately by Radial Basis Function neural network and Random Forest machine learners for performance comparison purposes. The simulation study showed that the proposed intrusion detection system offers high performance for detecting anomalies and normal activities with respect to a set of well-accepted metrics including detection rate, accuracy, and missed and false alarm rates.**

*Keywords*: **one-class classification, PCA, intrusion detection, host-based, system call trace data, random forest ensemble, radial basis function neural network.**

## I. Introduction

Intrusion Detection Systems (IDSs) is an active research field due to an acute and urgent need for cybersecurity techniques against constantly evolving attacks on the computing infrastructures globally. The cyber threats are categorized into two different types: known attacks for which a so-called signature is available and unknown attacks which were never seen before. Dealing with known attacks is much more straightforward since there are many references providing useful information about their traits and patterns. However, for unknown attacks there is no such signature: all that is known is that it does not belong to the normal mode of operation given the existing model, profile or signature of what constitutes "normal" or known attack signatures.

Machine learning techniques offer a host of options to process raw data for clustering or classification and can be employed effectively to detect intrusions. Intrusion detection systems for anomalous circumstances can be implemented by either semi-supervised techniques when labeled data for one class (usually normal class) for training exist and unsupervised learning for the case when there is no labeled data. According to [1], many factors make the anomaly detection procedure challenging. Defining an exact and precise boundary for the normal (target) class is one significant reason. In many domains, the normal behavior continually evolves as do the attacks and today's outlook often may not be a proper representation of future behavior. Moreover, in some cases, the anomalous tracepoint close to the normal boundary will be considered as a safe behavior. The adversaries that impose anomalous behavior on the system will try to adapt themselves to project a behavior as if it belonged to normal activity, which makes the detection process much more difficult. In most machine learning-based approaches for supervised classification, the availability of labeled data for both training and testing is necessary while for anomaly detection that is a major constraining issue.

There are numerous techniques reported in the literature for anomaly detection. Tavallaee *et al.* [2] compiled a comprehensive survey of anomaly-based intrusion detection studies. They have considered 276 articles in this area and reported that 160 of them employed classification-based methods; 62 of them proposed statistics-based techniques; 36 papers presented clustering approaches, and 46 studies utilized miscellaneous or hybrid methods. Another dimension for anomaly detection is the context for environment and execution: it can be based on host data, network traffic data or the combination of both host data and network traffic data. Following paragraphs will discuss recent work on host-based anomaly detection studies reported in the literature as that is the focus and scope of our study to be presented in this paper. Deshpande *et al*. [3] proposed a host-based anomaly-detection system model for cloud computing environment which alerts cloud users against intrusions within their system by analyzing the system call traces. In their study, the process ID, system calls, and their frequencies have been extracted from log records which were collected by the *Linux OS audit* framework. They stored the frequency of normal system calls into a "normal" records database. For any new system call trace converted to a frequency vector, their system compares it with the vectors in the "normal" database using the Euclidean distance. Utilizing the datasets used in [4] and [5], their study reports 96% accuracy in detecting malicious activities.

Distributed Denial of Service (DDOS) is one of the most common attacks in cloud environments. Although software-defined networking (SDN) provides various capabilities to prevent and mitigate DDoS attacks, they have several vulnerabilities [6] that can be exploited. Mahrach *et al.* [7] proposed a method to protect the cloud environment against SYN flooding attack at the switch level by detecting traffic anomalies and performing a SYN cookie technique. This study monitors the mean of a process based on the samples collected from that process during a particular period of time and triggers an alarm when the accumulated volume of measurements exceeds a threshold.

In a host-based misuse intrusion detection system study, Aghaei *et al.* [8] generated the most frequent unique *N*-gram features from system call traces of benign and different attack classes in the ADFA-LD dataset. To reduce the effect of noise, they also extracted the most frequent *N*-gram patterns regardless of their uniqueness according to their occurrence frequencies and resampled the training set using SMOTE technique to overcome the imbalanced data problem. Leveraging an ensemble learning model using five different classifiers including naïve Bayes, Support Vector Machine, PART, Decision Tree, and Random Forest, they created two different classification models. These were a binary classification model to detect attacks and a multiclass classification model to detect the type of attacks. They report 99.9% accuracy and 0.6% false positive rate for binary classification. However, the multiclass classifier performance was relatively poor as it failed to detect the attack type and only scored the average accuracy of 55% for all attack classes.

In another host-based misuse intrusion detection study on the ADFA-LD dataset, Serpen *et al.* [9], generated fixed-size feature vectors using a windowing technique. To reduce the dimensionality and to convert vectors in feature space into a set of linearly uncorrelated variables, they employed principal component analysis (PCA). To classify a new pattern, they leveraged the *k*-nearest neighbor algorithm using Euclidean distance to calculate its distance with the existing vectors generated during the training phase. They implemented both binary and multiclass classification models which had the same values for all performance metrics considered: their models achieved 99.9%, 0.2%, 100%, and 99.9% for accuracy, false positive rate, precision, and F1 score, respectively.

Against the backdrop of this comprehensive research work that is currently ongoing, there is still a need to explore intrusion detection systems for anomalous patterns and processes in order to increase performance to minimize false alarms and missed alarms, but more importantly, to maximize the detection rate for the anomalies. Consequently, in this study, we propose a host-based intrusion anomaly detection methodology, which draws upon techniques from principal component analysis, machine learning, and statistics. We employed a Linux-based dataset called ADFA-LD, which is a pool of system call traces during various normal and attack modes of operation [10-12]. This dataset contains thousands of traces during normal operation and six different attack types. The raw data of system call traces were preprocessed with the PCA-based feature extraction technique also called Eigentraces. Machine learning algorithms, Radial Basis Function (RBF) network and Random Forest, in conjunction with the Gaussian mixture models, were used to model probability distribution and density functions. Rendering a decision if a particular call trace pattern is anomalous or not employs the one-class classifier approach.

## II. Methodology

In this section, we define the overall data preparation and classifier model generation procedure. In subsequent sections, we present a windowing technique to generate equal-size vectors from the ADFA-LD raw dataset. This is followed by a PCA-based technique for feature extraction to generate the training and testing datasets. In the following section, we present the theoretical foundations of the one-class classification methodology that combines the density function and class probability estimation approaches. We then model target class through a user-defined distribution function and generate artificial data from that reference distribution, which is associated with the label for the second (anomaly) class. We take advantage of two machine learning algorithms, namely Radial Basis Function neural network and Random Forest, for target class probability estimation (but not for prediction). In the succeeding step, using Bayes' theorem, we estimate the target class density function, which is then utilized to decide if a newly encountered test pattern belongs to the target class or not.

### A. Pre-processing and feature extraction using windowing

The ADFA-LD dataset comprises three major categories of data, namely TDM, VDM, and ADM. TDM and VDM represent the normal class that contains 833 and 4372 system call traces, respectively. The ADM represents 746 attack traces. In order to generate the training set, we use the TDM only and for the testing set, we combine VDM and TDM.

The length of traces in ADFA-LD varies from 75 to 4494 system call instances. To apply the PCA-based template generation (aka the Eigentraces) methodology on this dataset, all traces must be equalized in length. One option is to define a window of size $d$, $d \in Z^+$, which slides over traces by a shift count or length $s$ where $s \leq d$. Thus, the windowing process generates trace files where their lengths are $d$ system call numbers. One drawback of this method is that the length of a trace might be incompatible with the shift size for the very last window of a trace file. In other words, the shift size might be greater than the remaining trace length. In this case, we can add a constant number which is a dummy or non-existent system call number as many times as needed to the end of the trace. Here is an illustrated example. Suppose there is a sequence of numbers with a length of 10 as below:

| $\alpha_1$ | $\alpha_2$ | $\alpha_3$ | $\alpha_4$ | $\alpha_5$ | $\alpha_6$ | $\alpha_7$ | $\alpha_8$ | $\alpha_9$ | $\alpha_{10}$ |
|---|---|---|---|---|---|---|---|---|---|

Consider sliding a window with size 6 ($d = 6$) to the right by shift length 5 ($s = 5$). Each rolled window forms an input pattern: accordingly, $\alpha_1$ to $\alpha_6$ is such an input pattern as shown below:

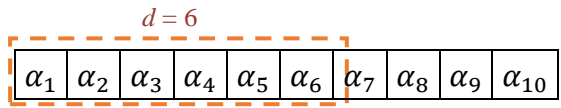

Now the window will shift to the right over this sequence by 5 numbers resulting in the pattern as below:

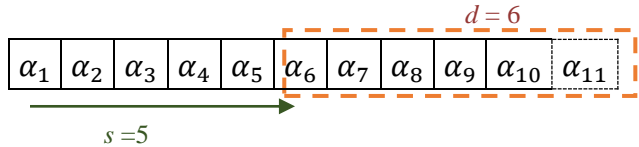

Note that it will be necessary to assign a number to any cells in the final window, which are blank such as $\alpha_{11}$. This needs to be done in a neutral manner. This number can be any value that is not a valid system call. For instance, if the range of numbers is between 1 and 100 for system calls for a given operating system, the number 0 can be added for the position $\alpha_{11}$. Another option is to throw away the very last window of system call numbers if the window is not complete

For ADFA-LD, we have assigned, through an empirical process of exploration, 76 to the window size with 10 as the shift length or size since the minimum trace file size is 75 system calls. In addition, we have used the value 0.1 to fill in the empty slots at the end of any window frame if the shift size is greater than the remaining length of a trace. This procedure is performed to convert the traces with different lengths into a collection or set of vectors of 76 elements, which are stored in the columns of a matrix associated with each of normal and attack classes. Since this is the anomaly detection context, the training set contains labels only for the target class, which is also called the normal class. The testing set includes patterns belonging to the normal class and the attack class. Table 1 shows the dimensions of training and testing matrices.

| Pattern Matrix | Intended Use | Pattern Class Membership | Dimensions |
|---|---|---|---|
| $\mathbf{D}_{training}$ | Training | Normal | 76×25689 |
| $\mathbf{D}_{test}$ | Testing | Normal/Attack | 76×195890 |

*Table 1.* Training and testing matrices.

### B. PCA-based feature extraction for a sequence of system calls

Principal Component Analysis (PCA) [13] uses a linear transformation to the maximal variation to find out principal components and to reduce the dimensionality of the data by identifying the direction of each principal vector, which is also called a principal component. The number of these components is much less than the number of variables. As a result, each original data instance vector can be represented by a smaller number of variables or features. This property makes the PCA a useful tool for analysis in high dimensional feature spaces.

To apply the PCA-based Eigentraces [14] methodology on the ADFA-LD system call trace data, we first form the data matrix using the trace vectors $\mathbf{t}_i$ with dimensions of 76×1,

where $i$=1,2,…, $M$ with $M$ representing the number of trace vectors:

$$\mathbf{D} = [\mathbf{t}_1 : \mathbf{t}_2 : \ldots : \mathbf{t}_M].$$

Figures 1 and 2 present the application of the Eigentraces methodology for training and testing procedures, respectively.

---

**Eigentraces Methodology: Generation of Training Template Set**

I. Form training data into a matrix for the normal class as follows. Let $\mathbf{t}_i$ represent the $i$th trace of the normal class where each trace is a 76×1 column vector of integers; then the training data matrix is defined as follows:

$$\mathbf{D}_{train} = [\mathbf{t}_1 : \mathbf{t}_2 : \ldots : \mathbf{t}_{25689}],$$

where there are *25689* traces belonging to the normal class. In other words, the training matrix has 25689 columns where each column represents a 76×1 trace (pattern) vector.

II. Compute the average trace vector **c** as follows:

$$\mathbf{c} = \frac{\mathbf{t}_1 + \mathbf{t}_2 + \ldots + \mathbf{t}_{25689}}{25689}$$

III. Compute the **P** matrix as follows:

$$\mathbf{P} = [\mathbf{t}_1 - \mathbf{c} : \mathbf{t}_2 - \mathbf{c} : \ldots : \mathbf{t}_{25689} - \mathbf{c}]$$

IV. Compute the covariance matrix **Q** as follows:

$$\mathbf{Q} = \mathbf{P}\mathbf{P}^{\mathbf{T}},$$

where **Q** has the dimensions of 76×76.

V. Compute 76 eigenvectors $\mathbf{e}_k$, for $k$=1,2,…,76, of the covariance matrix **Q** and form the **E** matrix for training data set using all 76 eigenvectors as follows:

$$\mathbf{E} = [\mathbf{e}_1 : \mathbf{e}_2 : \ldots : \mathbf{e}_{76}],$$

where **E** is a 76×76 matrix.

VI. Project all trace vectors $\mathbf{t}_i$, where $i$=1,2,…, 25689, in the training data set onto the Eigenspace $\mathbf{S}_{train} = \{\mathbf{z}_{i,train}\}$, which is a set of vectors, by computing its inner product with each of the eigenvectors in **E** as follows:

$$\mathbf{z}_{i,train} = \mathbf{E}^{\mathrm{T}} \mathbf{t}_i,$$

where $\mathbf{z}_{i,train}$ is a 76×1 training vector for $i$=1,2,…, 25689.

---

**Figure 1.** Eigentraces procedure for training set generation

The training Eigenspace forms the model for the target class. Evaluating the classification performance of this model is done through the testing set. ADFA-LD data set provides labeled normal and anomalous traces, which are preprocessed for feature extraction using the Eigentraces procedure. The procedure described in Figure 2 presents the formation and mapping of the testing set into the training Eigenspace.

---

**Eigentraces Methodology: Generation of Testing Template Set**

I. Form the normal and attack testing data into a matrix $\mathbf{D}_{test}$, where each trace is represented as a 76×1 vector $\mathbf{y}_{normal,i}$, $i$=1,2,…,185196 for normal traces and $\mathbf{y}_{attack,j}$, $j$=1,2,…,10694 for attack traces as follows:

$$\mathbf{D}_{test} = [\mathbf{y}_{normal,1} \vdots \mathbf{y}_{normal,2} \vdots \cdots \vdots \mathbf{y}_{normal,185196} \vdots \mathbf{y}_{attack,1} \vdots \mathbf{y}_{attack,2} \vdots \cdots \vdots \mathbf{y}_{attack,10694}].$$

II. Renumber the $\mathbf{D}_{test}$ components as below:

$$\mathbf{D}_{test} = [\mathbf{y}_1 \vdots \mathbf{y}_2 \vdots \cdots \vdots \mathbf{y}_{185196} \vdots \mathbf{y}_{185197} \vdots \cdots \vdots \mathbf{y}_{195890}],$$

where $\mathbf{y}_j$ for $j$=1,2,…,185196 represents normal vectors and for $j$=185197,185198,…,195890 represents attack vectors.

III. Compute the inner product of each test trace vector $\mathbf{y}_j$ with training eigenvector matrix $\mathbf{E}$. The resultant is a 76×1 vector in the Eigenspace given by

$$\mathbf{v}_{j,test} = \mathbf{E}^{\mathrm{T}}\mathbf{y}_j,$$

where $\mathbf{S}_{testing} = \{\mathbf{v}_{j,test}\}$, for $j$=1,2,…,195890, represents the projection of test vectors onto the Eigenspace.

**Figure 2.** Eigentraces procedure for testing set generation

---

### C. One-class classification (OCC)

In most classification problems, the training dataset covers all the classes. However, in some cases, the training set contains just a single class labeled as the *target*. In other words, the classifier must be induced using samples belonging to a single class although once the classifier is deployed, patterns or samples belonging to other classes which were not present in the training set may appear. This type of learning problem is called one-class classification (OCC) [15]. The one-class classification is also called as outlier or novelty detection since the procedure is used to distinguish between the normal (target) and abnormal (anomalous or outlier) data with respect to the distribution of the training data that exclusively contain normal (target) samples.

There are two main approaches for the OCC reported in the literature. Pearson *et al* [16] suggested a density estimation based approach, through which a statistical distribution is fitted to the target data. In this methodology, the data points with low-density value are considered as outliers and belonging to the abnormal class. The difficulty with this approach is identifying a proper distribution function for the given instances which need to be chosen empirically. The other technique for OCC is based on generating artificial data [17] which makes it possible to apply binary/multi-class classification to the OCC problem. Generating artificial data makes it possible to convert the OCC problem to the multi-class problem which means the artificial data play the role of the second (abnormal) class. The approach entails formulating a decision boundary around the target (normal) data such that any data points, which fall outside of this boundary represent the outliers. This decision boundary also can be defined by existing multi-class learning classifiers [18]. Both of these OCC techniques depend on the values of various parameters which are chosen empirically, and their performance is highly sensitive to these parameter values.

In this study, the approach to address the OCC problem entails employing the standard two-class classification by generating artificial data to form the second (abnormal or anomaly) class. The artificial data comes from a known reference distribution such as multi-variate distribution that can be estimated from the target class. The objective of learning associated with this method is to estimate class probability accurately rather than minimization of classification error. Accordingly, machine learners can be employed as probability modelers for calculating the class probability. Bagging, neural networks, support vector machines (SVM), and decision trees are among the choices for such modeling and are poised to yield good class probability estimation. The artificial data, which is associated with the second class label, need to be as close as possible to the target class data points to be able to fit the decision boundary around the target data as tightly as possible. One option adopts the Gaussian distribution as the reference distribution for the target class and then generates the artificial data from that. Thereafter, the target class probability function is computed using a machine learner, which is adapted for estimation, and finally, the target class density function is calculated using a derivation through the Bayes' rule [19].

The Bayes' rule estimates the probability of an event based on previous information. Formally stated, it computes the posterior probability of an event by taking prior probability and likelihood function into account as follows:

$$P(T|X) = \frac{P(X|T)P(T)}{P(X)} \tag{1}$$

where the term $P(T)$ stands for the prior probability that estimates the probability of event $T$ before the observed evidence $X$; the marginal likelihood $P(X)$ is the probability of each observed event; the likelihood $P(X|T)$ represents the probability of observing $X$ given $T$; and *the* $P(T|X)$ denotes the posterior probability, which is the probability of event $T$ given the evidence $X$.

As it is shown in [15], one can use the Bayes' theorem to compute the target class density function given target class prior probability, target class probability, and artificial class (reference) density. Redefinition of the parameters for Bayes' equation for the study herein for one-class classification is as follows:

$T$: target class

$A$: artificial class

$x$: instance

$P(T)$: the prior probability of observing a target class instance

$P(x)$: instance marginal likelihood

$P(T|x)$ : target class probability function

$P(x|A)$ : artificial class density function

$P(x|T)$ : target class density function

Since there are two classes, the probability of instance $x$ equals to observing an instance of either target or artificial data. Thus, we have the following:

$$P(T|x) = \frac{P(x|T)P(T)}{P(x|T)P(T) + P(x|A)P(A)} \qquad (2)$$

Next step is to solve Equation 2 for $P(X|T)$ that will be used for one-class classification and as the target class density function. Consequently, further algebraic manipulation of Equation 2 as detailed in [13] leads to the following form:

$$P(x|T) = \frac{(1 - P(T))\, P(T|x)}{P(T)\,(1 - P(T|x))}\; P(x|A) \qquad (3)$$

Equation 3 combines the artificial class density $P(x|A)$, class probability function $P(T|x)$, and target class prior probability $P(T)$ and therefore, facilitates the computation of the density function of the target class $P(x|T)$. In order to estimate the $P(x|A)$, we apply the Gaussian distribution to the target data with a mean value of 0 and the standard deviation value of 1. Hence, the probability density function of artificial data is given by

$$P(x|A) = \frac{1}{\sqrt{2\pi}\sigma}\exp\left[-\frac{\left(x - \mu_A\right)^2}{2\sigma^2}\right] = \frac{1}{\sqrt{2\pi}}\exp\left[-\frac{x^2}{2}\right] \qquad (4)$$

Next, we generate artificial data from this distribution where the size of this artificial data is user-defined. In order to have a balanced dataset, we generate the same amount of data which equals to target data set size. Therefore, the probability of $P(T)$ and $P(A)$ have the same value of 0.5. Accordingly, we simplify Equation 3 by eliminating $P(T)$ and $1 - P(T)$ as follows:

$$P(x|T) = \frac{P(T|x)}{(1 - P(T|x))}\; P(x|A) \qquad (5)$$

where $x$ is $\mathbf{z}_{i,train}$ for $i = 1,2,\dots,25689$. Equation 5 computes the target density function during the training procedure that yields the distribution and probability of the target instances. It is also necessary to define, during the training phase, a decision boundary around the target data by imposing a threshold for the density function which is called the target rejection rate (TRR). In order to classify a new test instance $\mathbf{v}_{test}$ using OCC, one can first compute its density value given $T$:

$$P(\mathbf{v}_{test}|T) = \frac{P(T|\mathbf{v}_{test})}{(1 - P(T|\mathbf{v}_{test'}))}\; P(\mathbf{v}_{test}|A) \qquad (6)$$

Next step is to compare the value of $P(\mathbf{v}_{test}|T)$ with the threshold value based on the TRR to make a classification decision. In the following section, we discuss the radial basis function neural network for probability density function estimation.

### D. Radial Basis Function Network (RBFN)

A Radial Basis Function Network (RBFN) is an artificial neural network with three layers as input, hidden and output [20]. The RBFN in this study is used to model the target class probability function. For this purpose, the Gaussian Mixture Model (GMM) is employed [21]. GMM computes a single mixture distribution by a weighted linear combination of existing probability density functions. In other words, the mixture distribution is a weighted average of the individual components. An RBFN uses the radial basis function as the activation function for its hidden layer neurons. A radial basis function, although typically Gaussian in form, can also be multiquadric, inverse quadric, inverse multiquadric, polyharmonic, or thin-plate spline. In this study, we employ the Gaussian function.

The dimensionality of the input patterns dictates the number of neurons in the input layer. The number of hidden layer neurons depends on the number of clusters in the input pattern space. The number of neurons in the output layer depends on the number of classes (labels) in the dataset for classification problems while, for function approximation, the RBFN has only one output layer neuron with a linear activation function. The Gaussian radial basis function has two parameters, namely center (mean) and variance. Equation 7 represents this function:

$$\phi_v = \exp\left(-\frac{(\mathbf{x} - \mu_v)^2}{2\sigma_v^2}\right) \qquad v \in \{1, 2, \dots, M\}, \qquad (7)$$

where the parameter $\mu$ is the center (vector); $\sigma$ refers to the standard deviation which is calculated using the training samples: and $(\mathbf{x} - \mu_v)^2$ is the distance between the input vector $\mathbf{x}$ and the $v$-th center given that there are $M$ centers. Suppose there are $K$ input vectors $\mathbf{x}$ where each one has $h$ features. In other words, $\mathbf{x}$ is an $h$-dimensional feature vector where each dimension is scalar-valued with $M << K$. A clustering algorithm can be used to generate $M$ clusters of the feature vectors. Subsequently, the variance can be calculated using Equation 8:

$$\sigma^2 = \frac{(maximum\ distance\ between\ any\ two\ centers)^2}{number\ of\ centers}. \qquad (8)$$

The mapping function $f$ for $i$-th neuron at the output layer is defined as follows:

$$f_i = \sum_{j=1}^{m} w_j\ \phi_j(\mathbf{x}) \qquad i \in \{1, 2, \dots, G\}, \qquad (9)$$

where there are $G$ neurons in the output layer.

RBFN training generates $M$ Gaussian distributions likely with different values for the two parameters, namely the center and the variance. Each of these distributions has a corresponding probability density function $\gamma_v(\mathbf{x}) = N(\mathbf{x}|\mu_v, \sigma_v^2)$. The resultant probability distribution is given by

$$\Psi(\mathbf{x}) = a_1 N(\mathbf{x}\mid \mu_1, \sigma_1^2) + a_2 N(\mathbf{x}\mid \mu_2, \sigma_2^2) + \dots +$$

$$a_M N(\mathbf{x}\mid \mu_M, \sigma_M^2) = \sum_{v=1}^{M} a_v N(\mathbf{x}\mid \mu_v, \sigma_v^2), \qquad (10)$$

where $\mu$ and $\sigma$ are the parameters of distributions, $M$ refers to the number of Gaussian distributions (clusters), and $a_i$, for $i = 1,2,\dots,M$, is the weight given to each individual component.

As a multivariate Gaussian model, each distribution has three different parameters: these are the input vector $\mathbf{x}$, the mean vector $\mu$, and the covariance matrix $\boldsymbol{\Sigma}$ where each principal diagonal element is the variance $\sigma^2$ of the random variable. In summary, if there are $M$ components (Gaussian distributions) and $\phi_v$ represents the component $v$, general GMM assumption is as follows:

- Each component $\phi_v$ has a mean vector $\mathbf{\mu}_v$.
- Each component $\phi_v$ generates an instance from corresponding Gaussian distribution with mean $\mathbf{\mu}_v$ and covariance matrix $\Sigma_v$.

Considering given parameters, the GMM probability density function is defined as

$$\Psi(\mathbf{x}) = \sum_{v=1}^{M} P(\phi_v)N(\mathbf{x} \mid \mathbf{\mu}_v, \Sigma_v), \qquad (11)$$

where

$$P(\phi_v) > 0, P(\phi_1) + P(\phi_2) + \cdots + P(\phi_M) = 1,$$

$$\mathbf{\mu}_v = \frac{1}{K_v} \sum_{i=1}^{K_v} \mathbf{x}_i, \text{ and}$$

$$\Sigma_v = \frac{1}{K_v} \sum_{i=1}^{K_v} (\mathbf{x}_i - \mathbf{\mu}_v)(\mathbf{x}_i - \mathbf{\mu}_v)^T,$$

where $K_v$ refers the total number of input vectors in cluster $v$ where $v = 1, 2, \ldots, M$.

Next step is to generate a cluster with a defined probability function using Expectation Maximization (EM) [22]. The EM algorithm has two steps. In step one, for each data point $\mathbf{x}_i$, it computes the probability that it belongs to the component $\phi_v$. Then it estimates the posterior probability under $\phi_v$ using Bayes' rule:

$$P(\phi_v|\mathbf{x}_i) = \frac{P(\phi_v)P(\mathbf{x}_i|\phi_v)}{\sum_{b=1}^{m} P(\phi_b)P(\mathbf{x}_i|\phi_b)} \qquad (12)$$

where $P(\mathbf{x}_i|\phi_v) = N(\mathbf{x}|\mathbf{\mu}_v, \Sigma_v)$, $P(\mathbf{x}_i|\phi_b) = N(\mathbf{x}|\mathbf{\mu}_b, \Sigma_b)$ and the terms in the denominator must sum to 1.0. In the second step, for each component, GMM updates its parameters using the weighted data points. Both steps will be repeated until convergence. Further details on various processes to optimize the parameter values are provided in [22].

Consequently, an RBFN models input training data using Gaussian distributions through its hidden layer neurons; the GMM algorithm defines a mixture distribution of the entire training data, and the EM algorithm computes the target probability function for use in the proposed one-class classification framework.

### E. Random Forest

Random Forest inductive learning algorithm [23] is an ensemble design [24-26] which employs decision trees as base learners and the divide-and-conquer approach to implement a high-performing learner. Given the set $S$ with $|S|$ data points $\mathbf{x}$ (instances) where $\mathbf{x}$ is an $h$-dimensional vector, divide $|S|$ data points into $C$ number of random subsets where each subset is used to induce one decision tree. Each decision tree independently makes a prediction and final prediction is based on the majority vote among all decision tree predictions.

The procedure for the application of the random forest algorithm for training and testing is as follows:

1) Generate $C$ subtrees from training set $S$ by iterating Bootstrap Aggregation (Bagging) $C$ times. Bagging methodology generates these subtrees with replacement by uniformly sampling from $S$. Since, the replacement is allowed, some data samples might be used to induce one or more subtrees.

2) Apply Random Subspace creation (attribute bagging) to generated subtrees. If each data point $\mathbf{x}_i$ described by $h$ features, attribute bagging randomly selects $h' < h$ attributes and learns the decision tree [27].

3) The attribute, which results in the best split according to an objective function, will be used to do the split on that node. The $h'$ value is empirically set, however, Breiman [28] suggested three values as follows: $0.5\sqrt{h}$, $\sqrt{h}$ and $2\sqrt{h}$.

4) Repeat the same procedure with different $h'$ attributes: the iteration count value is user-defined.

5) Each decision tree predicts the class of an instance independently and the consensus class, which is predicted the most often among the base learners, becomes the final decision.

For inducing a decision tree as a base learner for Random Forest ensemble for a binary-class problem with two values as "positive" and "negative", the root node is assigned the best attribute (see [29] how to choose the best attribute), which contains all instances. The process of tree induction continues by splitting the node for all possible attribute values and including entire instances with that attribute value in the corresponding child nodes. If all instances associated with a node result in positive (negative) class value, then it is called a pure node, where the tree development stops. As the leaves of a decision tree must be pure, splitting the non-pure node for its attribute values takes place again, which is followed by checking to see if any of the nodes are pure. At some point during the decision tree development as a base learner, if there is no attribute left and the leaf node has both positive and negative instances, the procedure for decision tree development must be repeated anew with different attributes assigned to the root and child nodes.

The class probability estimates for a given input pattern can be calculated based on the frequency of observed class values occurring at the leaves. The probability of class T being responsible for generating the instance $\mathbf{x}$ in the subtree $S_g$ is calculated by the number of leaves resulting in class T divided by the total number of leaves:

$$P_g(T|\mathbf{x}) = \frac{\text{Number of Leaves with Class Label T in } S_g}{\text{Total Number of Leaves in } S_g},$$

where $\mathbf{x}$ belongs to the union of training and testing sets, namely $\mathbf{S}_{training} = \{\mathbf{z}_{i,training}\}$ for $i = 1,2,\ldots,25689$ and $\mathbf{S}_{testing} = \{\mathbf{v}_{j,test}\}$, for $i = 1,2,\ldots,195890$.

In terms of an ensemble of several decision trees, one can use relative class frequency to define the class probability distribution by computing the average of the relative class frequencies of the forest sub-trees:

$$P_{Forest}(T|\mathbf{x}) = \frac{1}{C} \sum_{g=1}^{C} P_g(T|\mathbf{x}) \qquad (14)$$

Equation 14 yields the target class probability function.

*F. Threshold value computation*

In order to classify a new test instance $\mathbf{v}_{j,test}$, it is necessary first to assign a probability threshold for values computed using the estimated density function $P(\mathbf{x}|T)$ during the training procedure. Rejection sampling or acceptance-rejection method [30] is a type of Monte Carlo method that generates observations from a known distribution: it generates a proposal distribution from a given density function and performs uniformly random sampling from that proposed distribution. Thereafter, it rejects samples, which are not under the coverage of the original density function. We use the standard Gaussian distribution as the proposal distribution and run the rejection sampling process with given training vectors $\mathbf{x}$ and corresponding target class density function $P(\mathbf{x}|T)$, where $\mathbf{x}$ represents the set of vectors defined by $\mathbf{z}_{i,training}$ for $i = 1, 2, \ldots, 25689$ in this study. For a range of threshold values between 0.001 and 0.4, the Area Under the Curve (AUC) metric is calculated using both RBFN and Random Forest algorithms for estimating the target class probability function as it is shown in Figure 3. The same figure shows AUC values are ascending for the true rejection rate (TRR) value range from 0.001 to 0.05 and descending for higher TRR values. Therefore, 0.05 is the optimum threshold value for probability density function based on the AUC values, which means any higher or lower value results in lower true positive rate or higher false negative rate. Thus, for any new instance $\mathbf{v}_{j,test}$, it will be considered as an outlier if its value computed through the estimated probability density function given target class $P(\mathbf{v}_{j,test}|T)$ is less than 0.05. However, this is a data-driven threshold which might vary across different datasets and should be estimated during the training process.

## III. Simulation Study

We used MATLAB™ R2016a for data preprocessing and implementation of the Eigentraces methodology, and WEKA 3.9 one-class classifier package for classification. We considered both the RBFN and Random Forest as probability estimators for two separate studies. For RBFN, we generated 5 clusters and assigned 0.1 as the minimum standard deviation of each cluster. For the Random Forest algorithm, the number of execution slots to use for constructing the ensemble is set as 1, and the number of randomly chosen attributes equals to $\log_2 76 \approx 6$. The value 76 refers to the number of attributes in our dataset. For the one-class classification toolset in Weka, Table 2 presents the user-settable parameters and their values. Classification performance measurements have been done through the CRAN package repository provided for the R programming framework.

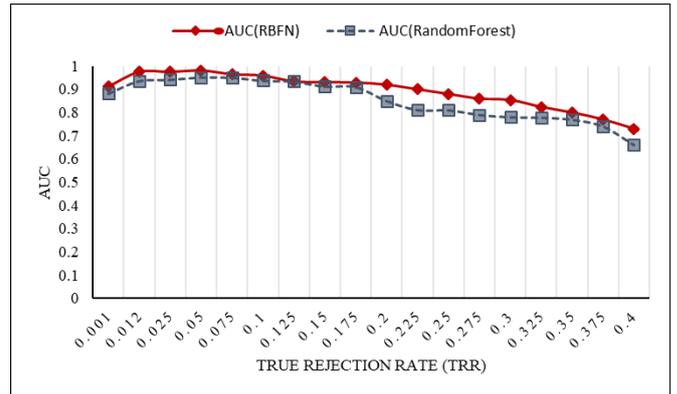

**Figure 3.** A trade-off between threshold and AUC values: the highest AUC value belongs to the threshold value of 0.05

| OCC Parameter | Value |
|---|---|
| *numRepeates* | 10 |
| *numericGenerator* | GMM |
| *mean* | 0.0 |
| *standardDeviation* | 1.0 |
| *percentageHeldout* | 10 |
| *proportionGenerated* | 0.5 |
| *targetRejectionRate* | 0.05 |

*Table 2.* WEKA one-class classifier parameter values

Figures 4 and 5 present a summary highlight of training and testing procedures.

*A. Classifier development*

We used Gaussian density with a diagonal covariance matrix containing the observed variance of each attribute in the target (normal) data on the training templates to define the reference distribution. Using the reference distribution, we generated 25689 artificial data points which collectively belong to artificial data set $A$ and calculated $P(\mathbf{z}_{i,training}|A)$. Then, we used two different algorithms separately, namely RBFN and Random Forest, to estimate the target class probability function. Finally, we combined the artificial class density function and target class probability function to calculate the target class density function with 10-fold cross-validation for validating the proposed function with the target rejection rate of 0.05.

The accuracy of the decision boundary in this single class prediction is evaluated by the True Positive Rate (TPR), defined as the ratio of the number of correctly classified instances to the number of total instances of the target class. In other words, these values represent the rates for correctly classified target instances. AUC values as presented in Table 3 are 98% for the RBFN and 95.1% for the Random Forest, which are high enough to employ in the testing process.

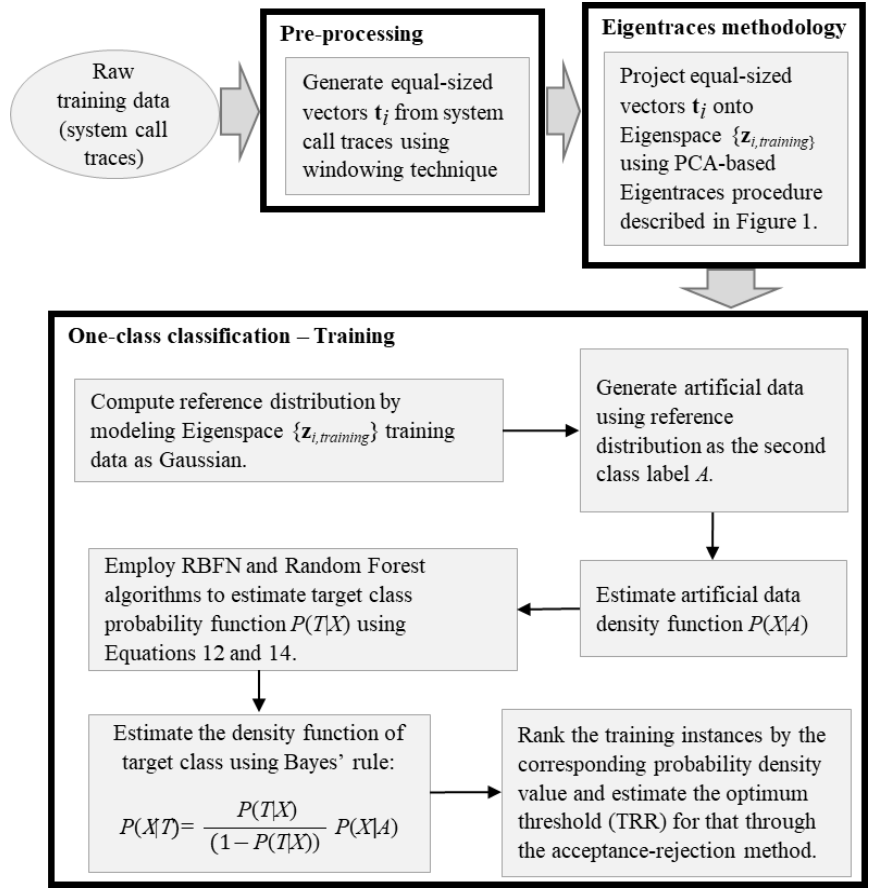

**Figure 4.** Training procedure flowchart

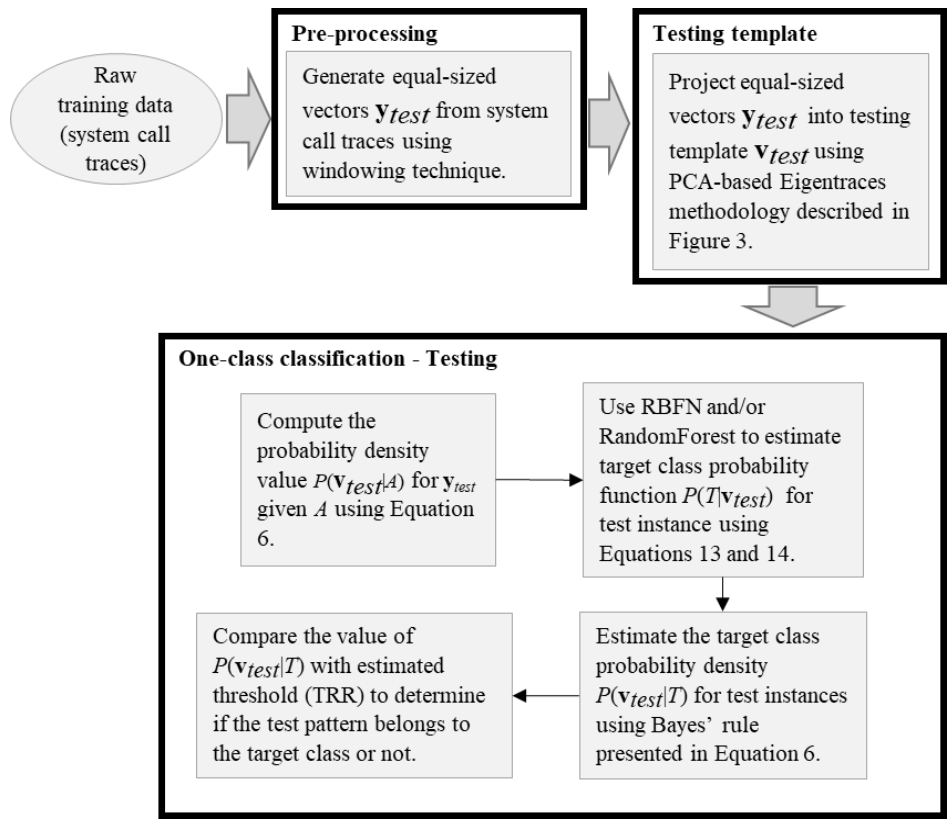

**Figure 5.** Testing procedure flowchart

| Probability Modeler | RBFN | Random Forest |
|---|---|---|
| **No of Instances** | 25689 | 25689 |
| **Correctly Classified** | 25150 | 24433 |
| **Misclassified** | 539 | 1256 |
| **TPR** | 0.979 | 0.951 |
| **FPR(FAR)** | 0.02 | 0.049 |
| **AUC** | 0.98 | 0.951 |

*Table 3.* Performance evaluation for the single target class

### B. Performance for unseen data

We next evaluate the performance of the generated model by applying an external testing set, which consists of instances belonging to both normal and attack classes. The density function value of target class for each test instance, $P(\mathbf{v}_{test}|T)$, is calculated and compared with the threshold that is computed empirically during the training phase. If the value of $P(\mathbf{v}_{test}|T)$ is less than or equal to 0.05, the test instance $\mathbf{v}_{test}$ will be considered as an anomaly; otherwise, it belongs to the target or normal class. We report the following four evaluation metrics to show the performance of the proposed method:

$$Detection\ Rate\ (DR) = \frac{True\ Negatives}{True\ Negatives + False\ Positives}$$

$$False\ Positive\ Rate\ (FPR) = \frac{False\ Positives}{False\ Positives + True\ Negatives}$$

$$False\ Alarm\ Rate\ (FAR) = \frac{False\ Positive\ Rate\ (FPR) + False\ Negative\ Rate\ (FNR)}{2}$$

$$Accuracy = \frac{True\ Positive\ Rate\ (TPR) + True\ Negative\ Rate\ (TNR)}{Total\ Number\ of\ Test\ Instances}$$

Table 4 shows the values of all four performance metrics while Table 5 presents the confusion matrices for both RBFN and Random Forest cases. As both of these tables indicate, the proposed method exhibits excellent performance. The missed alarm rates are low as only 43 attacks are considered as normal for the RBFN and 71 attacks are considered normal for the Random Forest algorithm.

### C. Anomaly detection studies reported on ADFA-LD data set in the literature

Literature abounds with studies about different types of two-class anomaly-based intrusion detection systems for a variety of contexts such as host-based, network-based, and application-based and for various datasets. A comparison with those studies that employed the ADFA-LD dataset for anomaly detection is presented next.

Chawla *et al.*[31] proposed a combined convolutional neural network (CNN) and recurrent neural network (RNN) model as an anomaly detection system using the ADFA-LD dataset. In their study, CNN layers capture the local correlations of structures in the system call traces and RNN layer learns the sequential correlations from the features. This model is trained on normal traces and predicted the probability distribution for the next system call in each trace. Accordingly, it predicts the probability for the entire trace, and imposing a threshold based on the range of the negative log-likelihood, it classifies new traces into normal or abnormal class. This study reports the attack detection rate of 100% with a very high 60% false alarm rate.

| Probability Modeler | RBFN | Random Forest |
|---|---|---|
| **No of Instances** | Normal = 185196 Attack = 10694 Total = 195890 | |
| **Correctly Classified** | 194217 | 193689 |
| **Misclassified** | 1673 | 2201 |
| **DR** | 0.996 | 0.993 |
| **FPR** | 0.004 | 0.006 |
| **FAR** | 0.006 | 0.009 |
| **Accuracy** | 0.996 | 0.993 |

*Table 4.* Model evaluation detailed performance metrics values

| Probability Modeler: RBFN | | Predicted As | |
|---|---|---|---|
| | | *Normal* | *Attack* |
| **True Class** | *Normal* | 183566 | 1630 |
| | *Attack* | 43 | 10651 |

| Probability Modeler: Random Forest | | Predicted As | |
|---|---|---|---|
| | | *Normal* | *Attack* |
| **True Class** | *Normal* | 183066 | 2130 |
| | *Attack* | 71 | 10623 |

*Table 5.* Confusion matrices of for class probability estimators: RBFN and Random Forest

Xie, et al. [32] used a short sequence method for feature extraction and formulation on the ADFA-LD dataset and applied one-class support vector machine (SVM) for multiclass classification. They reported 80% accuracy with false positive rate (FPR) of 15%. Using a frequency-based multiclass model, Xie, et al. [33] reduced the dimension of the dataset. They applied kNN and k-Means Clustering (kMC) algorithms for the detection of anomalies. Their main conclusion was that the kNN or KMC algorithms were not promising to detect anomaly attacks as their kMC implementation resulted in 60% accuracy with 20% FPR.

Doyle III [34] created a frequency-based two-class model using *N*-grams method and leveraged the kNN and SVM to perform the classification and reported 60% accuracy. Haider et.al. employed the zero-watermark algorithm in two different studies. In their first study [35], they proposed a character data zero-watermark inspired statistical-based feature extraction strategy for integer data. They evaluated the performance of RBF kernel, SVM, and kNN classifiers on the binary class version of the anomaly detection problem. They reported 78% detection rate (DR) with 21% false alarm rate (FAR).

All anomaly detection studies resulted in intrusion detection systems with lower-than-acceptable performance profiles. Table 5 represents a comparison between our study and others reported in the literature on the ADFA-LD dataset for anomaly detection.

| Study | Accuracy | FPR | DR | FAR |
|-------|----------|-----|-----|-----|
| Chawla *et al.*[31] | - | - | 1.00 | 0.60 |
| Xie, et al. [32] | 0.80 | 0.15 | - | - |
| Xie, et al. [33] | 0.60 | 0.20 | - | - |
| Doyle III [34] | 0.60 | - | - | - |
| Haider, et.al. [35] | - | - | 0.78 | 0.21 |
| **RBFN** | **0.99** | **0.004** | **0.99** | **0.006** |
| **Random Forest** | **0.99** | **0.006** | **0.99** | **0.009** |

*Table 6.* Comparison with other anomaly detection studies on ADFA-LD dataset

## IV. Conclusion

In this study, a semi-supervised host-based anomaly detection approach using one-class classification on ADFA-LD dataset has been proposed. A windowing method was implemented in order to generate equal-sized vectors of system call traces. A PCA-based methodology on the windowed trace data was employed to generate the trace templates or Eigentraces. Probabilistic modeling of the training data was accomplished through the Gaussian Mixtures and Expectation-Maximization methods in conjunction with the Radial Basis Function Neural Network and Random Forest machine learners. One-class classification through 10-fold cross-validation was used for evaluating the anomaly detection performance. Performance evaluation employed a testing set which contained a large number of normal and anomalous instances. The proposed method demonstrated very high performance with respect to a multitude of benchmark metrics and comparative to other studies as reported in the literature on the same data set for anomaly detection.

## Author Biographies

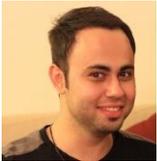

**Ehsan Aghaei**  He was born in Tehran, IRAN in 1991. He obtained his B.Sc. in Computer Engineering from the University of Shahid Beheshti, Tehran, IRAN, and received his M.Sc. in Computer Science from the University of Toledo, OH. Ehsan is currently a Computer Science Ph.D. candidate at the University of North Carolina-Charlotte working at 'Cyber Defense and Network Assurability Research Center (CyberDNA)' and NSF Cybersecurity Analytics and Automation (CCAA)' center as a research assistant. His research domain primarily pertains to the area of Machine Learning, Deep Learning, Text Mining, and NLP for Intrusion Detection, Security Analytics, and Cyber Automation

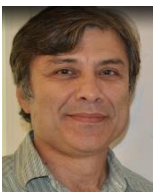

**Dr. Gursel Serpen**  Dr. Serpen received a Ph. D. in Electrical Engineering (with specialization in computer engineering) from the Old Dominion University, Norfolk, Virginia in 1992. He worked as an application and senior software engineer for Integrated Systems, Inc. (acquired by WindRiver Systems, Inc in late 90s) of Santa Clara, California between 1992 and 1993. He joined the Computer Science and Engineering Department at the University of Toledo as a faculty member in 1993, and has been serving as a faculty member with the Electrical Engineering and Computer Science Department at the University of Toledo to date. His main research interests entail Artificial Intelligence and Machine Learning with applications to problems in a variety of problem domains including automation, optimization, robotics, bio-medical informatics, and cyber-security.